# Benchmarking Time Series Forecasting Models: From Statistical Techniques to Foundation Models in Real-World Applications

Issar Arab
Ruhrdot GmbH
Munich, Germany
Email: issar.arab@tum.de

Rodrigo Benitez
Ruhrdot GmbH
Berlin, Germany
Email: rodrigobenitezdev@gmail.com

*Abstract*— Time series forecasting is essential for operational intelligence in the hospitality industry, and particularly challenging in large-scale, distributed systems. This study evaluates the performance of statistical, machine learning (ML), deep learning, and foundation models in forecasting hourly sales over a 14-day horizon using real-world data from a network of thousands of restaurants across Germany. The forecasting solution includes features such as weather conditions, calendar events, and time-of-day patterns. Results demonstrate the strong performance of ML-based meta-models and highlight the emerging potential of foundation models like Chronos and TimesFM, which deliver competitive performance with minimal feature engineering, leveraging only the pre-trained model (zero-shot inference). Additionally, a hybrid PySpark-Pandas approach proves to be a robust solution for achieving horizontal scalability in large-scale deployments.

## I. INTRODUCTION

AI and machine learning have shown remarkable potential in solving complex problems across diverse domains. They have revolutionized fields such as natural language processing [1], protein sequence analysis [2][3][4], mass spectrometry proteomics [5], drug discovery [6][7][8], software testing and fault localization [9], and the wide field of time series [10]. Time series forecasting has evolved significantly, moving from classical statistical approaches like ARIMA [11] to advanced machine learning and deep learning techniques [12] such as RNNs, LSTMs [13], and transformers [14][15]. Recent advancements in foundation models introduce promising avenues for multivariate and univariate time series forecasting with minimal feature engineering.

This study addresses the challenge of forecasting hourly sales for thousands of restaurants across Germany over the next 14 days. Using 2 years historical sales data, weather, and location-specific calendar features, the objective is to optimize staffing and procurement decisions based on the future sales volume on hourly basis in a restaurant. Using a sample of 4 restaurants located in different regions in the country, we benchmarked statistical, ML-based, conventional deep learning, and foundation models, analyzing their performance, scalability, and practicality for real-world deployment for large-scale production.

## II. METHODS

### A. Feature generation

To train our forecasting models, we compiled a set of features that expand the available information, providing the models with a comprehensive view of the various factors that could influence sales volume at a restaurant. These independent variables can be categorized into the following groups:

| Group | Feature |
| --- | --- |
| Weather-Related Features | is_rainy |
|  | is_snowy |
|  | is_clear |
|  | humidity |
|  | temperature |
| Time of Day Features | hour_sin |
|  | hour_cos |
| Day of the Week Features | day_of_week_sin |
|  | day_of_week_cos |
|  | is_monday |
|  | is_tuesday |
|  | is_wednesday |
|  | is_thursday |
|  | is_friday |
|  | is_saturday |
|  | is_sunday |
| Week of Year Features | week_of_year_sin |
|  | week_of_year_cos |
| Event/Activity Features | is_peak_hour |
|  | is_dinner |
|  | is_lunch |
|  | is_weekend |
| Holiday/Calendar Features | is_christmas_day |
|  | is_all_saints_day |
|  | is_ascension_day |
|  | is_ascension_day |
|  | is_whit_monday |
|  | is_corpus_christi |
|  | is_easter_monday |
|  | is_easter_sunday |
|  | is_epiphany |
|  | is_german_unity_day |
|  | is_good_friday |
|  | is_international_womens_day |
|  | is_labor_day |
|  | is_new_years_day |
|  | is_reformation_day |
|  | is_repentance_and_prayer_day |
|  | is_second_day_of_christmas |

## B. Forecast Models Analyzed

To conduct a thorough evaluation of various forecasting techniques, we benchmarked the following models across different categories:

| Category | Model | Library |
|---|---|---|
| Statistical Models | Prophet, multivariate Prophet | Prophet |
| Machine Learning | Linear Regression (Baseline), XGBoost, LightGBM | scikit-learn, MLForecast |
| Deep Learning | N-Beats [16] | N-Beats |
| Foundation Models | TimesFM [14] (multivariate transformer), Chronos [15] (univariate and multivariate transformers) | TimesFM [torch], AutoGluon |

This benchmarking encompasses a diverse set of models ranging from statistical approaches to cutting-edge foundation models, allowing for a comprehensive comparison. Regarding transformer-based models, we evaluated TimesFM and four sizes of the Chronos-Bolt pretrained foundation models (Tiny, Mini, Small, and Base). These four models were evaluated using zero-shot inference taking as input the time series data only, except for the "Chronos-Bolt (Base)" model, which was also assessed with the regressors option too.

## C. Evaluation Metrics

To assess the performance of the forecasting models, a set of standard evaluation metrics was employed. These metrics were selected to provide a holistic view of the models' performance, robustness, and overall predictive quality. Each metric captures a distinct aspect of model performance, enabling a well-rounded comparison. The metrics and their respective descriptions are outlined in the table below:

| Metric | Description | Desired Performance Trend |
|---|---|---|
| MSE | Mean Squared Error: Measures the average squared difference between actual and predicted values. | Lower |
| MAE | Mean Absolute Error: Measures the average absolute difference between actual and predicted values. | Lower |
| RMSE | Root Mean Squared Error: Square root of MSE, providing error in the same units as the target variable. | Lower |
| MAPE | Mean Absolute Percentage Error: Measures error as a percentage of actual values. | Lower |
| $R^2$ | Coefficient of Determination: Indicates the proportion of variance in the target explained by the model. | Higher |
| Corr | Correlation Coefficient: Measures the strength and direction of the linear relationship between actual and predicted values. | Higher |

## III. RESULTS AND DISCUSSION

We discuss the study from four different points: dataset structure, performance analysis, scalability, and finally recommendations.

### A. A Representative Diverse Dataset

The dataset in this study comprised time series data from four restaurants, each randomly selected from a different restaurant chain and geographically distributed. To maintain confidentiality, we refer to them as Restaurants A, B, C, and D. This selection ensured diversity in time series patterns, capturing various trends, seasonality, and unique operational dynamics. Forecasts were generated for the next 14 days for each restaurant, and the results were benchmarked across all models. By including restaurants with different patterns, the dataset provided a comprehensive evaluation of the forecasting models.

### B. Performance Analyisis

The performance analysis, as visualized in the figures 1, 2, 3, ,4, and 5, highlighted several key findings:
- Machine Learning Models:
  - ML-based models, particularly the ones implementing gradient boosting algorithms, consistently achieved the highest performance across all restaurants, leveraging Pandas Data Frames as input for training.
  - Scalability for Retraining: To enable accurate periodic retraining at scale, a hybrid Spark-Pandas solution was deployed. This approach combined Spark's scalability for distributed data processing with Pandas' flexibility for training, using user-defined functions (UDFs). This hybrid method demonstrated excellent stability, horizontal scalability, ease of implementation, and robust performance.
  - These features make ML-based solutions the most effective and practical choice for large-scale deployment.
- Foundation Models:
  - Chronos-Bolt foundation models provided competitive performance, matching the accuracy of ML-based models in 3 out of 4 cases.
  - Their zero-shot inference capability allowed them to capture trends, seasonality, and special patterns directly from the data without requiring additional regressors or extensive feature engineering.
  - While foundation models strike a good balance between computational complexity, performance, and feature engineering, they are GPU-dependent, which may limit their scalability in environments with constrained computational resources.

- Statistical and Deep Learning Models:
  o Prophet models, while effective in capturing seasonality, underperformed compared to both ML-based and foundation models.
  o The N-Beats deep learning model showed promise but required extensive computational resources and feature engineering, making it less efficient for this use case.

*C. Scalability*

- ML-Based Models:
  o The combination of Spark's scalability and Pandas' flexibility enables efficient horizontal scaling, making ML-based models well-suited for large-scale, distributed deployment.
  o This hybrid solution ensures accurate periodic retraining of models at scale, providing robustness and stability.
- Foundation Models:
  o Foundation models, such as Chronos-Bolt, scale by adding GPUs and require minimal preprocessing. This simplicity makes them an attractive option for environments with abundant GPU resources, but their dependency on GPUs—given that CPU execution yielded suboptimal results—could limit their widespread deployment in resource-constrained settings.

*D. Recommendations*

Based on the results and analysis, we propose the following recommendations:

- For Scalable Accurate Systems:
  o ML-based meta-models are the optimal choice for environments requiring highly scalable and robust solutions. The hybrid Spark-Pandas approach combines the best of both worlds, enabling efficient handling of large-scale data while maintaining high accuracy and stability—all without relying on GPU-enabled or dependent clusters.
- For Simplicity and Rapid Deployment:
  o Foundation models, particularly Chronos-Bolt, offer a compelling alternative for environments with abundant GPU resources. Their ability to provide accurate forecasting without requiring additional regressors or extensive feature engineering makes them ideal for simplified deployment pipelines.

## IV. CONCLUSION

This study highlights the potential of hybrid Spark-Pandas solutions and foundation models in large scale multi-time series forecasting. While ML-based models, particularly gradient boosting algorithms, remain a better choice for accuracy and robustness, pretrained foundation models represent a promising direction for scenarios prioritizing simplicity and minimal preprocessing (zero-shot inference). The results showcase a balanced approach between scalability, computational complexity, and performance, with the hybrid Spark-Pandas solution emerging as a practical and effective choice for large-scale, accurate deployment, particularly in the absence of advanced computational resources such as GPUs.

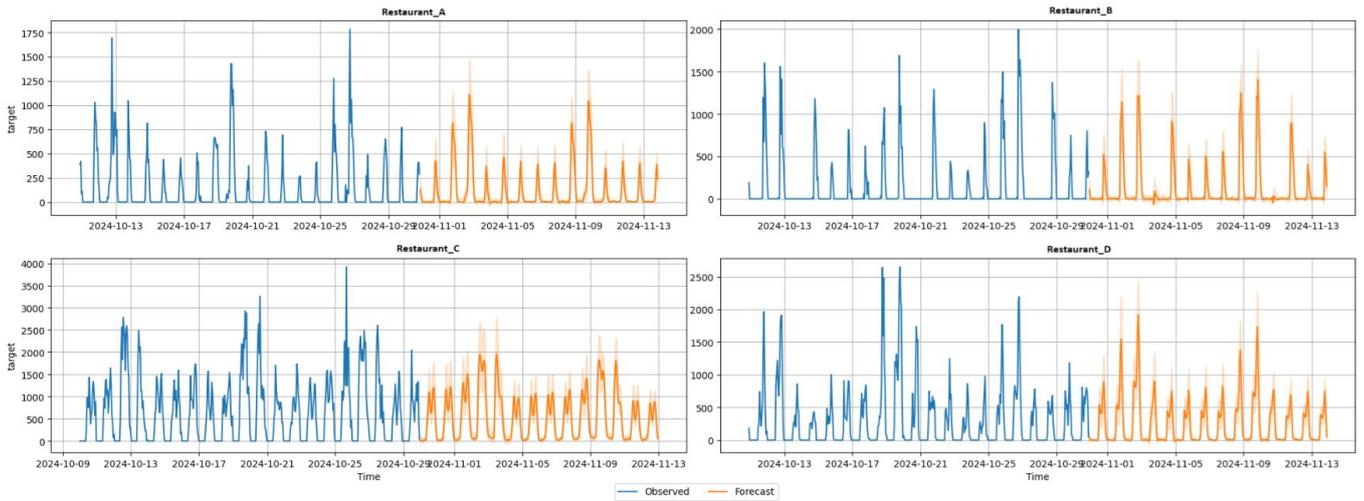

**Figure 1**: Chronos-Bolt (Base) accurate 14-day forecast using only time series data for all four restaurants.

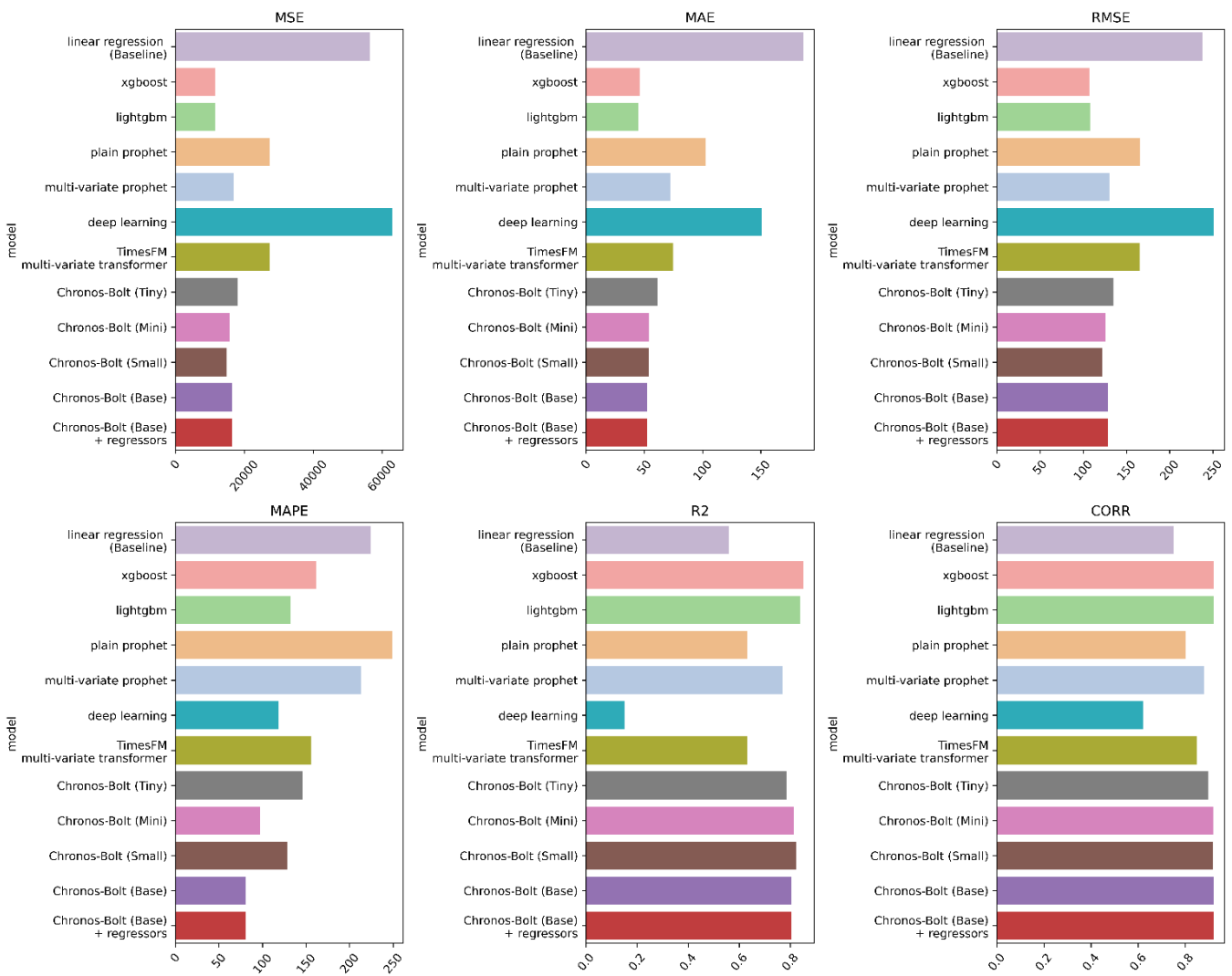

**Figure 2**: Benchmarking results of the 12 models evaluated for 14-day horizon forecasting in Restaurant A.

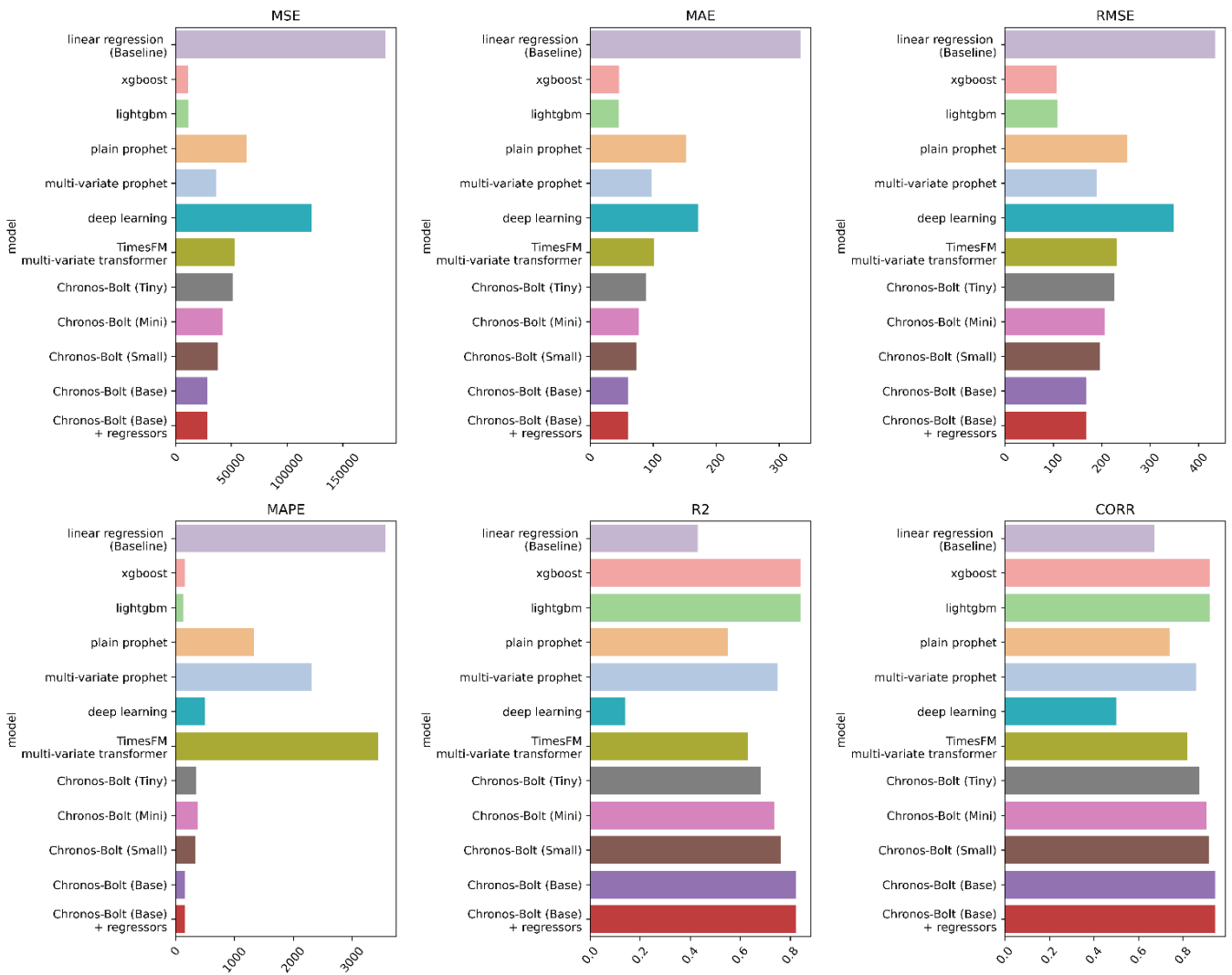

**Figure 3**: Benchmarking results of the 12 models evaluated for 14-day horizon forecasting in Restaurant B.

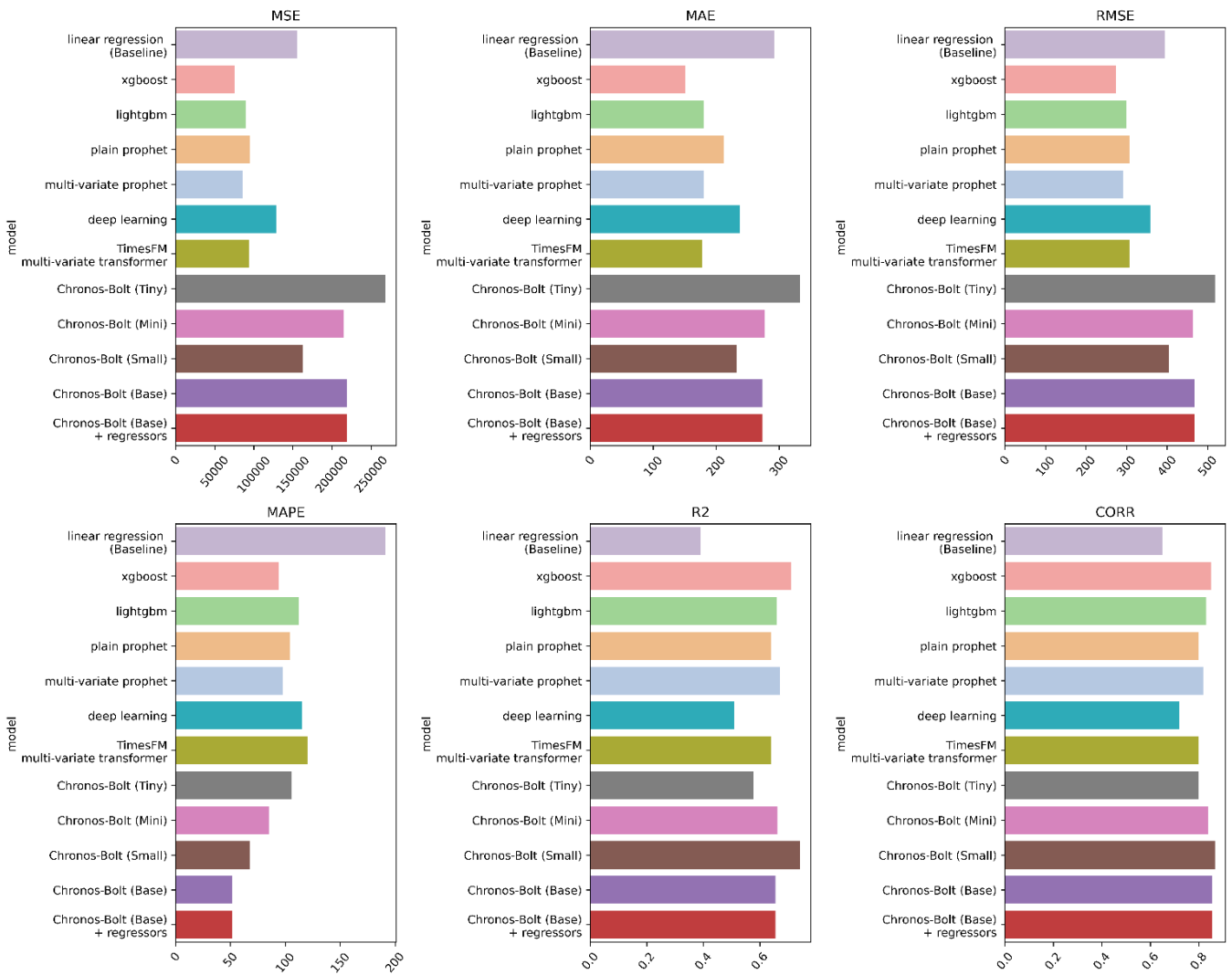

**Figure 4**: Benchmarking results of the 12 models evaluated for 14-day horizon forecasting in Restaurant C.

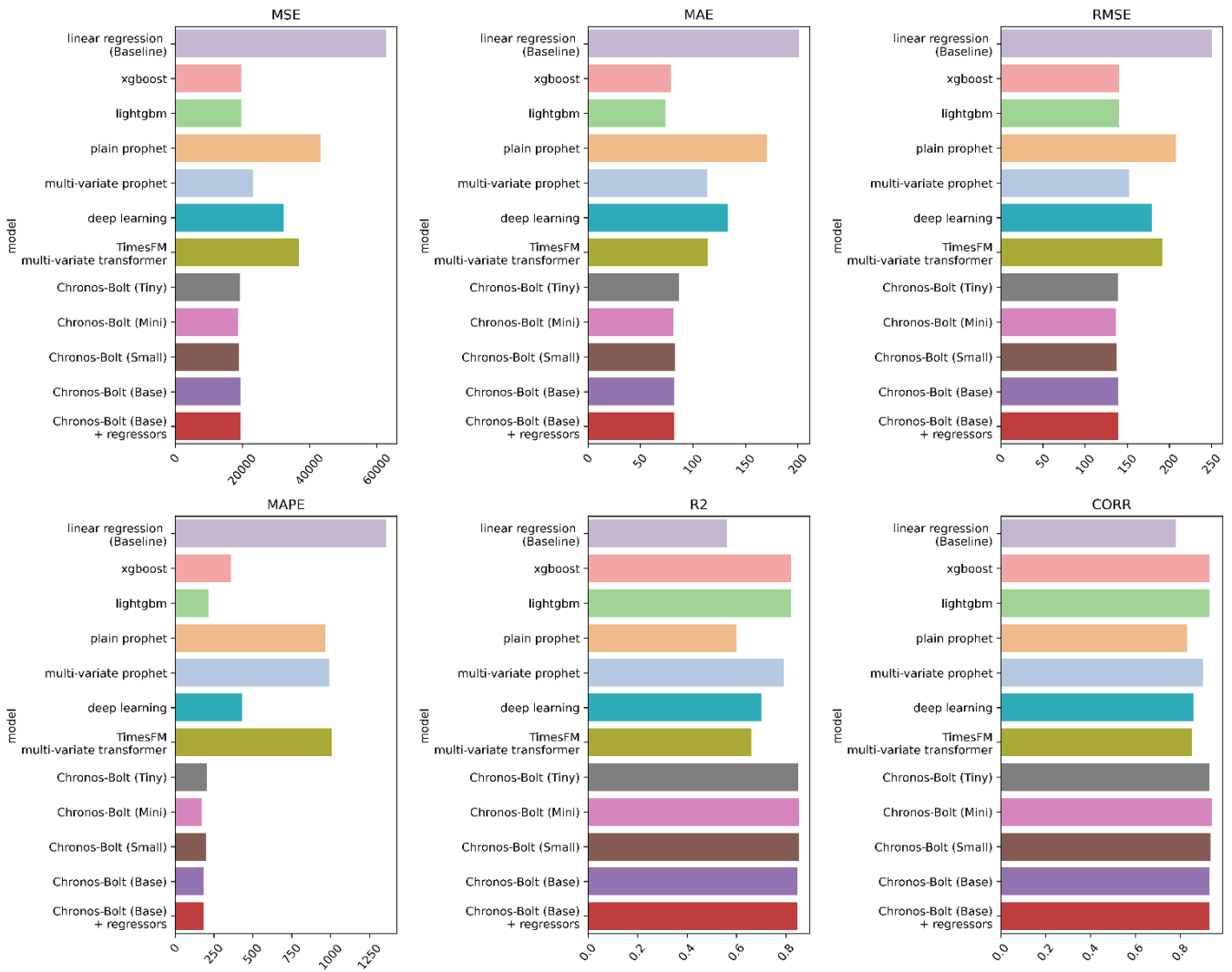

**Figure 5**: Benchmarking results of the 12 models evaluated for 14-day horizon forecasting in Restaurant D.